%% file: test.tex
\documentclass[runningheads]{llncs}

\usepackage[T1]{fontenc}
\usepackage{graphicx}
\usepackage{booktabs}
\usepackage{standalone}
\usepackage{hyperref}
\usepackage{array}
\usepackage{graphicx}
\usepackage{subcaption}
\usepackage{amsmath,amssymb}
\usepackage{url}
\usepackage{hyperref}
\usepackage{xcolor}
\usepackage{orcidlink}
\usepackage[nolist,nohyperlinks]{acronym}
\usetikzlibrary{positioning,fit,calc,arrows.meta}
\usepackage{fontawesome5}
\usepackage{pgfplots}
\pgfplotsset{compat=1.18}

\acrodef{WI}[WI]{Writer Identification}
\acrodef{HTG}[HTG]{Handwritten Text Generation}
\acrodef{GR-RNN}[GR-RNN]{Global-Context Residual Recurrent Neural Network}
\acrodef{GAN}[GAN]{Generative Adversarial Network}

\title{Generating Synthetic Data with Diffusion Models for Writer Identification}

\titlerunning{Synthetic Data with Diffusion Models for Writer Identification}

\author{Hamza Yazbeck \and
Marco Peer\orcidlink{0000-0001-6843-0830} \and
Andreas Fischer\orcidlink{0000-0003-0069-3436}}

\authorrunning{H. Yazbeck et al.}

\institute{AIBEX Group\\
University of Fribourg, Switzerland
}

\begin{document}
\maketitle

% ---------------------------------------------------------------
% Abstract
% ---------------------------------------------------------------
\begin{abstract}
Writer identification aims to determine the author of a handwritten sample based on individual writing characteristics. In this work, we investigate the use of diffusion-based handwritten text generation for data augmentation in offline word-level writer identification, which is particularly useful when lacking data. Specifically, we employ DiffusionPen to synthesize writer-conditioned handwriting samples and study their impact on writer identification performance in both contemporary and historical settings. Experiments are conducted on the CVL dataset and the historical BullingerDB dataset. Our results show that synthetic augmentation consistently improves writer identification accuracy on binarized images, yielding gains of 1.2 percentage points on CVL and 1.33 on Bullinger. Furthermore, we observe that grayscale image representations outperform alternative input representations across all experimental settings, for which synthetic data does not improve performance. We additionally demonstrate that synthetic data is particularly effective in low-resource scenarios, where the relative improvement increases as the amount of available real training data decreases.

\end{abstract}

% ---------------------------------------------------------------
\section{Introduction}
% ---------------------------------------------------------------

Handwriting is a distinctive biometrics of human motor behaviour, shaped by factors such as motor learning, cultural background, writing habits, and biomechanical characteristics \cite{fragnet}. As a result, each writer develops an individual writing style that remains observable even at the level of a very small amount of handwriting, e.g., an isolated word \cite{gr-rnn}. Secondly, handwriting is a valuable biometric modality for applications including forensic document analysis, historical manuscript research, and archival studies \cite{bullinger}. Automated \ac{WI} aims to assign a handwriting sample to one of several known writer identities \cite{peer_saghog_2024}. Our work focuses on offline \ac{WI} in a word-level setting, where the goal is to identify the writer from a single handwritten word image without considering the textual content.

\begin{figure}
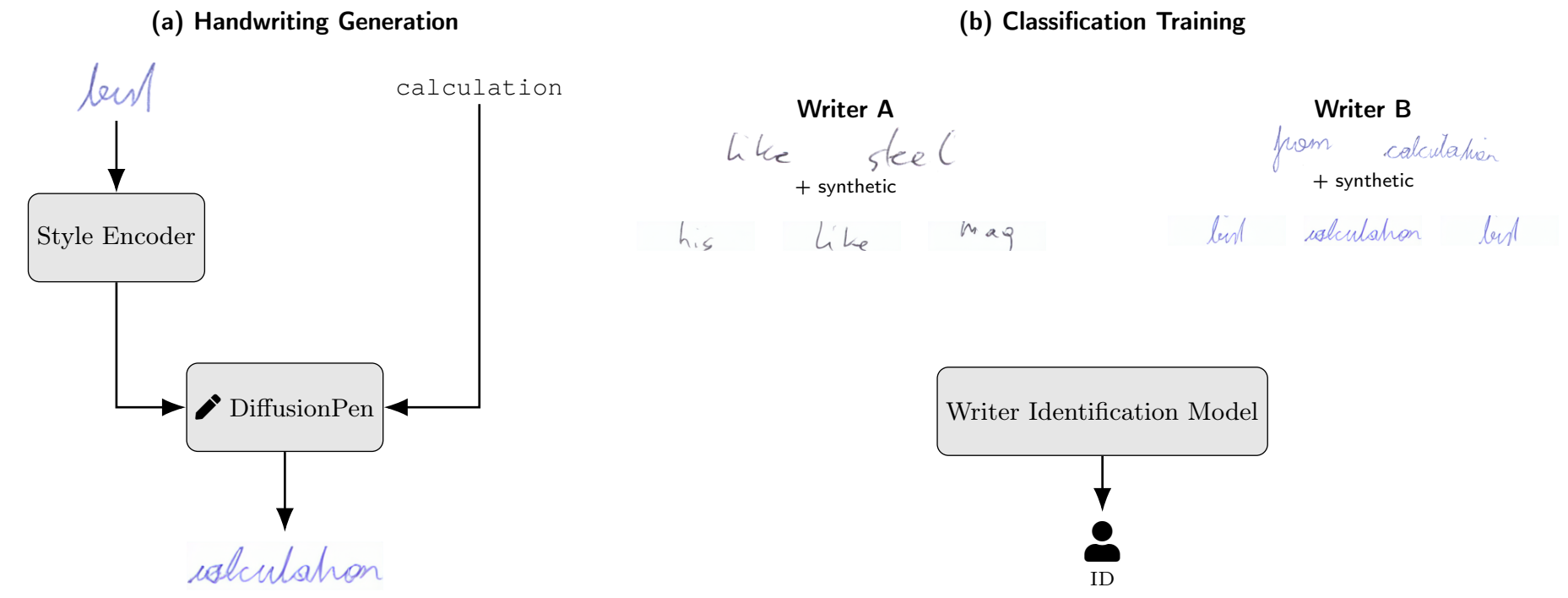

    \centering
    \includestandalone[width=\textwidth]{visual_abstract}
    \caption{Overview of our approach. We first generate handwriting with DiffusionPen \cite{diffpen} that is style-conditioned by forwarding images of a specific writer. Then, we use the synthetic data to augment the training process of our \ac{WI} model.}
    \label{fig:placeholder}
\end{figure}

In this paper, we investigate whether, and to what extent, \ac{HTG} can be used to improve the training of \ac{WI} models through data augmentation. Recent advances in diffusion-based generative models have improved the quality and stylistic consistency of synthesized handwriting \cite{wordstylist,diffpen}. In particular, prior work demonstrated that diffusion-generated handwriting preserves writer characteristics significantly better than earlier GAN-based approaches, achieving approximately 70\% writer identification accuracy compared to less than 5\% for GAN-generated samples \cite{wordstylist}. Motivated by these findings, we explore the use of diffusion-based handwriting synthesis to augment the training data for \ac{WI} across two distinct scenarios: (1) contemporary handwriting using the CVL dataset \cite{cvl}, and (2) historical handwriting using the BullingerDB dataset \cite{bullinger}. An overview of the proposed pipeline is illustrated in Fig.~\ref{fig:placeholder}. To the best of our knowledge, this is the first study investigating diffusion-based handwriting generation for historical documents in the context of writer identification.

Our experimental results demonstrate that synthetic augmentation improves writer identification performance in both settings, particularly when using binarized images. On the CVL dataset, we observe an improvement of 1.2 percentage points on Top-1 accuracy, while on the BullingerDB dataset we achieve 1.33. Furthermore, our experiments indicate that grayscale representations consistently outperform other image representations for \ac{WI}, both when training solely on real data and when incorporating synthetic samples. In addition, our findings confirm that synthetic data is particularly beneficial in low-resource scenarios. Similar to observations reported for text recognition \cite{muth2024}, we find that the relative benefit of synthetic augmentation increases when the amount of available real training data is reduced. Specifically, when using only 50\% of the original training data, synthetic augmentation yields an improvement of approximately 6\%.

The rest of our study is structured as follows: Section \ref{sec:rel_work} described related work, Section \ref{sec:method} provides details of our method, and Section \ref{sec:eval} shows the results of our approach. We conclude the paper in Section \ref{sec:conclusion}.

\section{Related Work}\label{sec:rel_work}
% ---------------------------------------------------------------

\paragraph{Writer Identification}

\ac{WI} is the task of assigning a handwriting sample to one of several known writer identities. In the closed-set setting, the writer of the query document is assumed to be present in the database (classification). In the open-set setting, the task is to find similar documents in a given gallery (retrieval). In this work, we focus on \ac{WI} and present related work in the following.

The first deep-learning-based approach for \ac{WI} is proposed by Fiel and Sablatnig, in which they use a CNN to identify the writer based on a single line image. The output of the penultimate layer is then used as a feature vector for pooling. This simple method already achieves accuracies of 98.9 \% on the CVL \cite{cvl} dataset.

Further research mostly focuses on word-level identfication. He and Schomaker propose FragNet \cite{fragnet}, an architecture for word-based \ac{WI} that is aimed to explore all writing style
information contained in the whole word image by adding a second path which takes fragments of the word image as input. Their follow-up work which this paper is based on is \ac{GR-RNN} \cite{gr-rnn} extends FragNet by adding gated recurrent units. We explain \ac{GR-RNN} in Section \ref{sec:method}.

Koepf et al. \cite{koepf} investigate the use of vision transformers for the \ac{WI} task on page-level and find that their application is limited due to the small dataset sizes (CVL \cite{cvl}, ICDAR 2013 \cite{icdar13}).

Recently, \ac{WI} has been investigated in the CircleID competition at ICDAR2026 \cite{gorges2026circleid} for which the writer of a hand-drawn circle has to be determined, with the additional task of recognizing foreign writers as outliers. \ac{WI} is also closely aligned with the open-set task of \emph{writer retrieval}, for which current research focuses on historical documents \cite{peer_saghog_2024}.

% The approach this paper is based on 
% is a deep architecture designed for writer identtification. It relies on the observation that local handwriting fragments can be highly informative for discriminating writers. The method extracts overlapping square patches from handwriting images, processes each patch independently using a CNN, and aggregates the resulting patch-level embeddings into a fixed-dimensional representation. This makes the approach suitable for documents of variable length and provides implicit data augmentation through patch extraction.

% \subsubsection{GR-RNN}
% The \ac{GR-RNN} \cite{gr-rnn} extends fragment-based writer identification by incorporating sequential modelling. A writer's style is not only reflected in isolated strokes but also in rhythmic and structural patterns across sequences of characters. GR-RNN processes fragments using a bidirectional recurrent module and combines the sequential representation with global context information through a residual connection.

\paragraph{Handwriting Synthesis} \acp{GAN} have been widely used for offline handwriting generation \cite{scrabblegan}. Early methods condition generation on text representations and often rely on auxiliary recognisers to improve readability. Later approaches, such as GANwriting \cite{GANwriting}, jointly condition on text and style references to generate images in seen and unseen writer styles. 

Although the handwriting is visually plausible, the style preservation lacks compared to the current dominant architecture of diffusion-based models. The first work is WordStylist, that also applied \ac{WI} as a evaluation task and show that diffusion models outperform \acp{GAN} by more than 60\% accuracy. The follow-up work DiffusionPen \cite{diffpen} is used for generating the handwriting in this work, and we briefly explain it in our methodology.

Another diffusion-based approach is proposed by Mayr et al. \cite{mayr} that targets paragraph-level generation.

% ---------------------------------------------------------------
\section{Methodology}\label{sec:method}
% ---------------------------------------------------------------

Our method includes two parts: First, we train DiffusionPen \cite{diffpen} as shown in Fig. \ref{fig:diffusionpen} to generate synthetic word images. Then, we apply a \ac{WI} model and use the synthetic handwriting to augment the training process. We describe the details in the following.

\subsection{Diffusion Model}

\begin{figure}[t]
    \centering
    \includegraphics[width=0.8\textwidth]{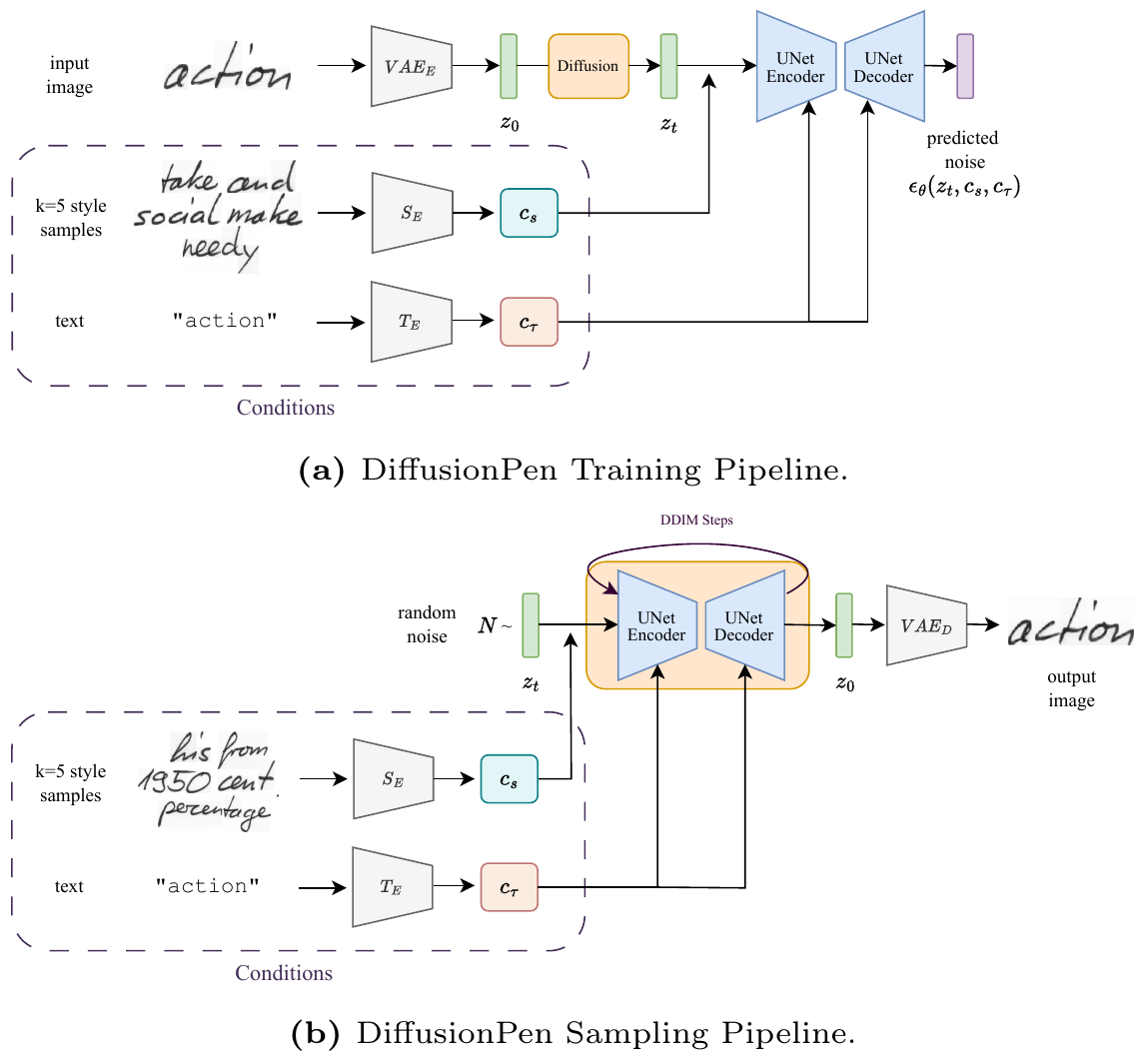}
    \caption{Overview of DiffusionPen \cite{diffpen}. The model combines a text encoder, a style encoder, and a conditional latent diffusion generator. During training, images are encoded into the latent space; during sampling, generated latents are decoded into handwriting images (Figure taken from \cite{diffpen} via \href{https://www.ecva.net/papers/eccv_2024/papers_ECCV/papers/11492.pdf}{open access version}).}
    \label{fig:diffusionpen}
\end{figure}

% Paragraph about what diffusion models are
DiffusionPen \cite{diffpen} is a latent diffusion model for few-shot styled handwriting generation. It generates word-level handwriting images conditioned on a text prompt and a small set of reference images from the target writer. The model consists of a latent autoencoder, a character-level text encoder, and a hybrid style encoder. An overview is shown in Fig. \ref{fig:diffusionpen}. The same text encoder as in the original implementation is used (Canine-C). The training process for applying DiffusionPen consists of two stages: style encoder training and denoising model training. 

The style encoder is trained to produce compact writer representation from reference images. We use the triplet loss as an objective, which aims to push embeddings from different writers further then embeddings of the same writer.

% Given two writer embeddings $e_A$ and $e_B$, DiffusionPen supports style interpolation through
% \begin{equation}
%     e_{\mathrm{mix}} = (1 - w)e_A + w e_B,
% \end{equation}
% where $w$ controls the interpolation weight.

The text encoder maps the target word into a character-level representation. During generation, the denoising network is conditioned on both the style embedding and text embedding. The diffusion model is trained to predict the noise added to a latent representation at timestep $t$ by minimising
\begin{equation}
    \left\| \epsilon - \epsilon_\theta(z_t, c_s, c_\tau) \right\|^2,
\end{equation}
where $z_t$ is the noisy latent, $c_s$ is the style condition, and $c_\tau$ is the text condition.

% TODO: What is DDIM?
% TODO: Confirm exact batch sizes and number of epochs for each dataset.
% TODO: Confirm whether all models were fine-tuned from the IAM DiffusionPen checkpoint.

\begin{figure*}[t]
    \centering

    % ===================== Row 1 =====================
    \begin{subfigure}{0.32\textwidth}
        \centering
        \caption*{Writer 2463}
        \includegraphics[width=\linewidth]{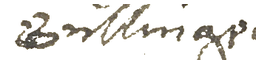}
    \end{subfigure}
    \hfill
    \begin{subfigure}{0.32\textwidth}
        \centering
        \caption*{Writer 8042}

        \includegraphics[width=\linewidth]{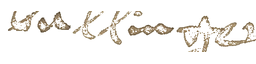}
    \end{subfigure}
    \hfill
    \begin{subfigure}{0.32\textwidth}
        \centering
        \caption*{Writer 8283}

        \includegraphics[width=\linewidth]{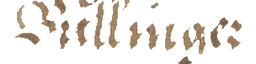}
    \end{subfigure}

    \vspace{0.5em}

    % ===================== Row 2 =====================
    \begin{subfigure}{0.32\textwidth}
        \centering
        \includegraphics[width=\linewidth]{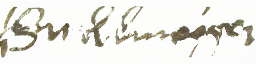}
    \end{subfigure}
    \hfill
    \begin{subfigure}{0.32\textwidth}
        \centering
        \includegraphics[width=\linewidth]{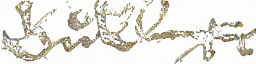}
    \end{subfigure}
    \hfill
    \begin{subfigure}{0.32\textwidth}
        \centering
        \includegraphics[width=\linewidth]{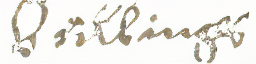}
    \end{subfigure}

    \vspace{0.5em}

    % ===================== Row 3 =====================
    \begin{subfigure}{0.32\textwidth}
        \centering
        \includegraphics[width=\linewidth]{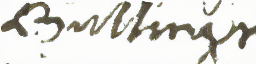}
    \end{subfigure}
    \hfill
    \begin{subfigure}{0.32\textwidth}
        \centering
        \includegraphics[width=\linewidth]{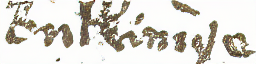}
    \end{subfigure}
    \hfill
    \begin{subfigure}{0.32\textwidth}
        \centering
        \includegraphics[width=\linewidth]{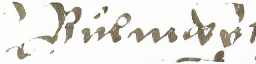}
    \end{subfigure}

    \vspace{0.5em}

    % ===================== Row 4 =====================
    \begin{subfigure}{0.32\textwidth}
        \centering
        \includegraphics[width=\linewidth]{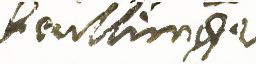}
    \end{subfigure}
    \hfill
    \begin{subfigure}{0.32\textwidth}
        \centering
        \includegraphics[width=\linewidth]{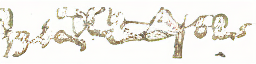}
    \end{subfigure}
    \hfill
    \begin{subfigure}{0.32\textwidth}
        \centering
        \includegraphics[width=\linewidth]{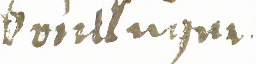}
    \end{subfigure}

    \caption{Real handwriting samples (top row) and generated samples (rows 2--4) for three different writers.}
    \label{fig:bullinger_samples}
\end{figure*}

\subsection{Synthetic Data Generation}

After training DiffusionPen, synthetic word images are generated for each writer. For every generated sample, five reference word images from the same writer are randomly selected to define the writer style. The target transcription is sampled from another writer's word image, excluding transcriptions that already appear for the target writer. This encourages textual diversity while keeping generated words within the vocabulary distribution of the dataset. Examples of the generated samples for the BullingerDB \cite{bullinger} dataset are shown in Fig. \ref{fig:bullinger_samples}. 

The generated images are then used to train the \ac{WI} model in two settings: synthetic-only training and mixed training, where synthetic images are added to the real training set. Performance is compared against a real-data-only baseline.

\subsection{Writer Identification Model}

We adopt the \ac{GR-RNN}, proposed in \cite{gr-rnn}, as our \ac{WI} model, as it shows state-of-the-art performance on word-level for contemporary datasets such as CVL \cite{cvl}. GR-RNN integrates global-context features with local fragment-level information extracted from CNN feature maps to represent handwritten word images. A gated recurrent unit is employed to capture contextual dependencies across the sequence of local fragments, while a residual RNN ensures that each fragment's representation retains both the between-fragment context and its original feature. The hidden state of the recurrent network is initialized with a global-context vector obtained via global average pooling of the final convolutional layer. The network takes a single word image as input and classifies the corresponding writer. We use GR-RNN following the original implementation as proposed by the authors.

\subsection{Datasets} 

Experiments are conducted on the CVL \cite{cvl} dataset consisting of contemporary handwriting and a subset of the historical BullingerDB \cite{bullinger} dataset. The dataset statistics are shown in Table~\ref{tab:datasets}.

% which subset?

\begin{table}[t]
\centering
\caption{Statistics of the datasets used in the experiments.}
\label{tab:datasets}
\begin{tabular}{lcccc}
\toprule
{Dataset} & {Writers} & {Total} & {Train} & {Test} \\
\midrule
CVL \cite{cvl} & 310 & 99,904 & 79,790 & 20,114 \\
BullingerDB \cite{bullinger} (Subset) & 596 & 452,640 & 362,114 & 90,562 \\
\bottomrule
\end{tabular}
\end{table}

The CVL dataset contains 310 writers and 99,904 word images. The BullingerDB dataset contains 796 writers and 4,427,128 word images. We construct a smaller subset by removing the 200 writers with the largest number of images, resulting in 596 writers and 452,640 word images (a writer has on average 600 word instances). Note that the BullingerDB corpus is challenging, in particular for generating syntethtic data, as the writers contributed over a longer period of time, meaning that the writing style may vary for an individual. 

For both datasets, 80\% of each writer's images are used for training and 20\% for testing.

% ---------------------------------------------------------------
\section{Evaluation}\label{sec:eval}
% ---------------------------------------------------------------
In the following, we describe our experimental setup and the main results of our paper. While the results are presented on the CVL benchmark, we end the section with a study on the BullingerDB dataset.

\subsection{Experimental Setup}

All word images are resized to $256 \times 64$ pixels.
 The model is evaluated on the fixed real test set of each dataset. We run experiments with three different seeds and the standard deviation across runs is less than 0.1 percentage points.

\paragraph{Training setup} WI model is trained for 50 epochs using the Adam optimizer with a learning rate of $10^{-4}$, a weight decay of $10^{-4}$, and a mini-batch size of 16. The learning rate is reduced by a factor of 0.5 every 10 epochs.

The style encoder is trained for 100 epochs using Adam with a learning rate of $10^{-3}$ and weight decay of $10^{-4}$. The learning rate is reduced by a factor of 0.1 every three epochs. 

The diffusion model is then trained using the frozen style encoder to provide the writer-style condition. We use a DDIM scheduler for noise injection and sampling. During training, the diffusion process uses 500 timesteps. During sampling, the number of reverse steps is reduced to 50. The denoising model is optimised with AdamW using a learning rate of $10^{-4}$ and weight decay of 0.2.

\paragraph{Data Generation} Synthetic data volume is expressed relative to the size of the real training set. For example, 100\% generated data means that the number of generated images per writer equals the number of real training images for that writer.

\subsection{Style Preservation}

First, we check the style preservation of DiffusionPen for the given writers. We take a GR-RNN trained on real CVL data and use it to classify the generated images, i.e., a train-on-real, test-on-synthetic protocol. Table~\ref{tab:stylepres} reports the results for both image representations, with the generated set sized at 100\% and 200\% of the real training data. We notice a strong performance gap between the two formats. We assume that the grayscale model partly relies on ink and pen cues that the generator, rendering in generic ink, does not reproduce, whereas the binarised model relies on stroke-level style, which the synthetic images do preserve. We use this as motivation for the format-dependent behaviour observed throughout the remaining experiments.

\begin{table}[htbp]
\centering
\caption{Style preservation on CVL. GR-RNN trained on real data classifies the generated images.}
\label{tab:stylepres}
\begin{tabular}{llcc}
\toprule
Image type & Generated ratio & Top-1 & Top-5 \\
\midrule
Grayscale & 100\% & 9.17  & 22.67 \\
Grayscale & 200\% & 5.03  & 16.05 \\
Binarised & 100\% & 78.19 & 91.85 \\
Binarised & 200\% & 78.08 & 91.68 \\
\bottomrule
\end{tabular}
\end{table}

\subsection{Effect of Image Format}

We focus on \ac{WI} performance on the CVL dataset under different configurations. We investigate three factors: the image format used to train DiffusionPen, the input format used for GR-RNN, and the amount of synthetic data added to the real training set. Table~\ref{tab:format_results} reports the Top-1 accuracy for different combinations of DiffusionPen training and GR-RNN input format. Three training conditions are compared: generated data only, real data only, and real data combined with 100\% generated data.

\begin{table}[t]
\centering
\caption{Top-1 accuracy (in \%) on CVL writer identification test set for different DiffusionPen image formats and GR-RNN input formats.}
\label{tab:format_results}
\begin{tabular}{llccc}
\toprule
DiffusionPen & GR-RNN & Syn. only & Real Only & Syn. + Real \\
\midrule
RGB & Grayscale & 66.76 & \textbf{94.29} & 94.09 \\
RGB & Binarised & 73.78 & 82.87 & \textbf{83.98} \\
Grayscale & Grayscale & 49.84 & \textbf{94.29} & 89.96 \\
Grayscale & Binarised & 50.75 & \textbf{82.87} & 80.23 \\
Binarised & Grayscale & 59.56 & \textbf{87.17} & 84.47 \\
Binarised & Binarised & 40.50 & 82.87 & \textbf{83.65} \\
\bottomrule
\end{tabular}
\end{table}

The results show that synthetic data improves performance when added to real data using binarized data. The real-only baseline for binarised GR-RNN is 82.87\%, independent of DiffusionPen's training format, since it does not involve synthetic data. The strongest improvement is observed when DiffusionPen is trained on RGB images and GR-RNN on binarised images, where Top-1 accuracy increases from 82.87\% to 83.98\%. In contrast, synthetic data generated by the grayscale DiffusionPen model reduces performance in both evaluated GR-RNN input formats, suggesting that not all generated data is equally beneficial. The state-of-the-art performance of GR-RNN is also achieved on grayscale images (92.6\% \cite{gr-rnn}). We believe that networks trained on RGB and grayscale images infer features from ink patterns or even different pens from the word images instead of the handwriting style \cite{peer_saghog_2024}. 

\subsection{Effect of Synthetic Data Volume}

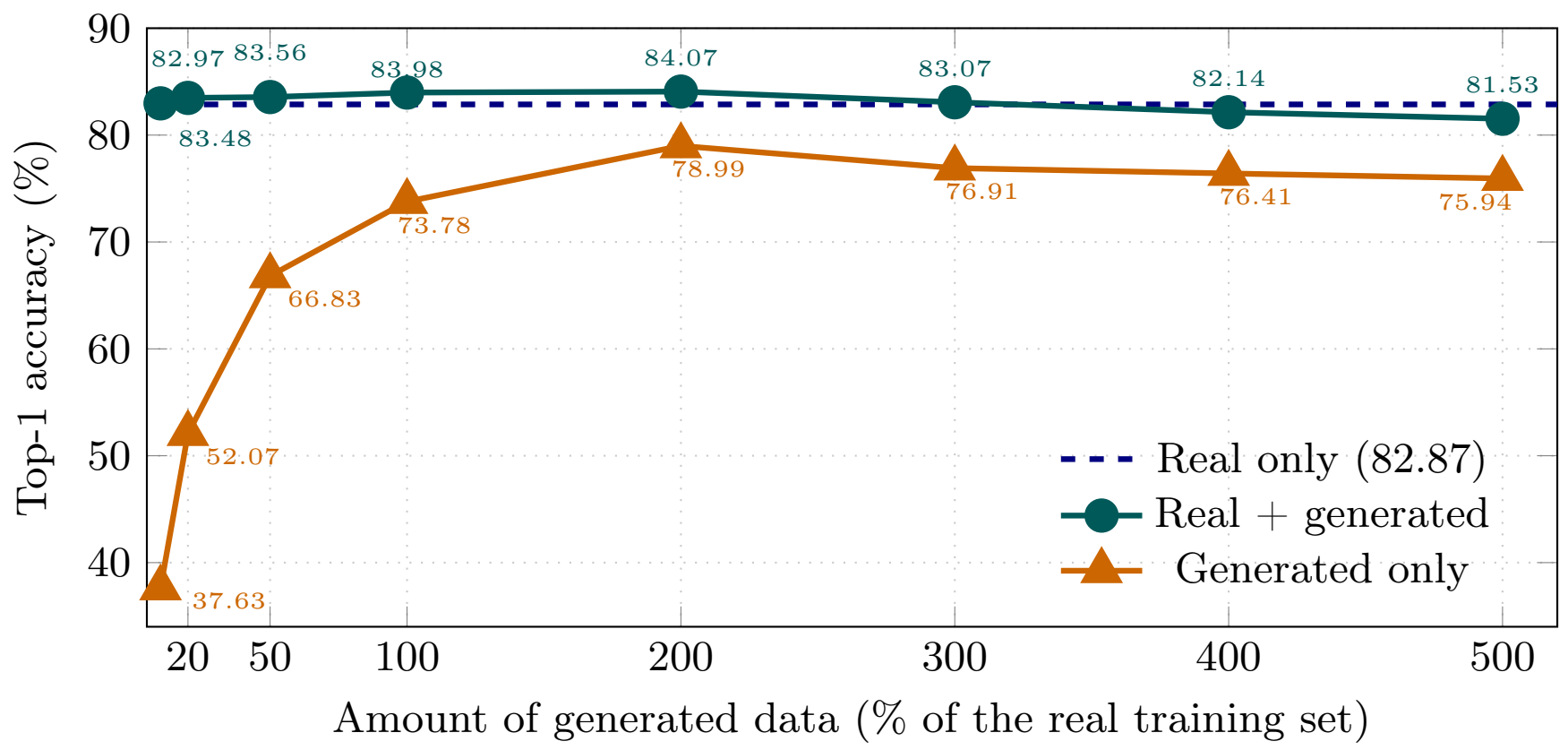
\begin{figure}[t]
    \centering
    \input{imgs/generated_data_plot}
    \caption{Top-1 accuracy on the CVL test set (binarised) as the amount of generated data grows from 10\% to 500\% of the real training set.}
    \label{fig:volume_curve}
\end{figure}

Table~\ref{tab:volume_results} reports Top-1 and Top-5 accuracy for RGB-trained DiffusionPen under different synthetic data volumes. Adding synthetic data slightly improve Top-5 accuracy for grayscale images, while we also observe that for binarized word images, the mixed scenario outperforms the networks trained on real data only. Using only synthetic data, the networks trail by approximately 28\% (grayscale) and 10\% (binarized). Fig.~\ref{fig:volume_curve} additionally tracks Top-1 accuracy as the amount of generated data is raised from 10\% to 500\% of the real training set using binarised images. Both conditions improve up to around 200\% and then decline, with the augmented condition eventually falling back below the real-only baseline. Beyond a moderate multiple of the real data, additional synthetic samples begin to dilute rather than reinforce the real signal, so there is a sweet spot rather than a monotone gain.

\begin{table}[t]
\centering
\caption{Effect of synthetic data volume on CVL writer identification accuracy (in \%) using RGB-trained DiffusionPen for the GR-RNN model.}
\label{tab:volume_results}
\begin{tabular}{llccc}
\toprule
{Training Data} & \multicolumn{2}{c}{{Grayscale}} & \multicolumn{2}{c}{{Binarised}} \\
\cmidrule(lr){2-3}\cmidrule(lr){4-5}
 & {Top-1} & {Top-5} & {Top-1} & {Top-5} \\
\midrule
Real Only & \textbf{94.29} & 98.66 & 82.87 & 93.98 \\
100\% Syn. + Real & 94.09 & \textbf{98.70} & 83.98 & 94.65 \\
200\% Syn. + Real & 93.97 & 98.67 & \textbf{84.07} & \textbf{94.97} \\
100\% Syn. Only & 66.76 & 84.37 & 73.78 & 89.08 \\
200\% Syn. Only & 66.57 & 84.43 & 78.99 & 92.12 \\
\bottomrule
\end{tabular}
\end{table}

\subsection{Effect of Real Data Availability}

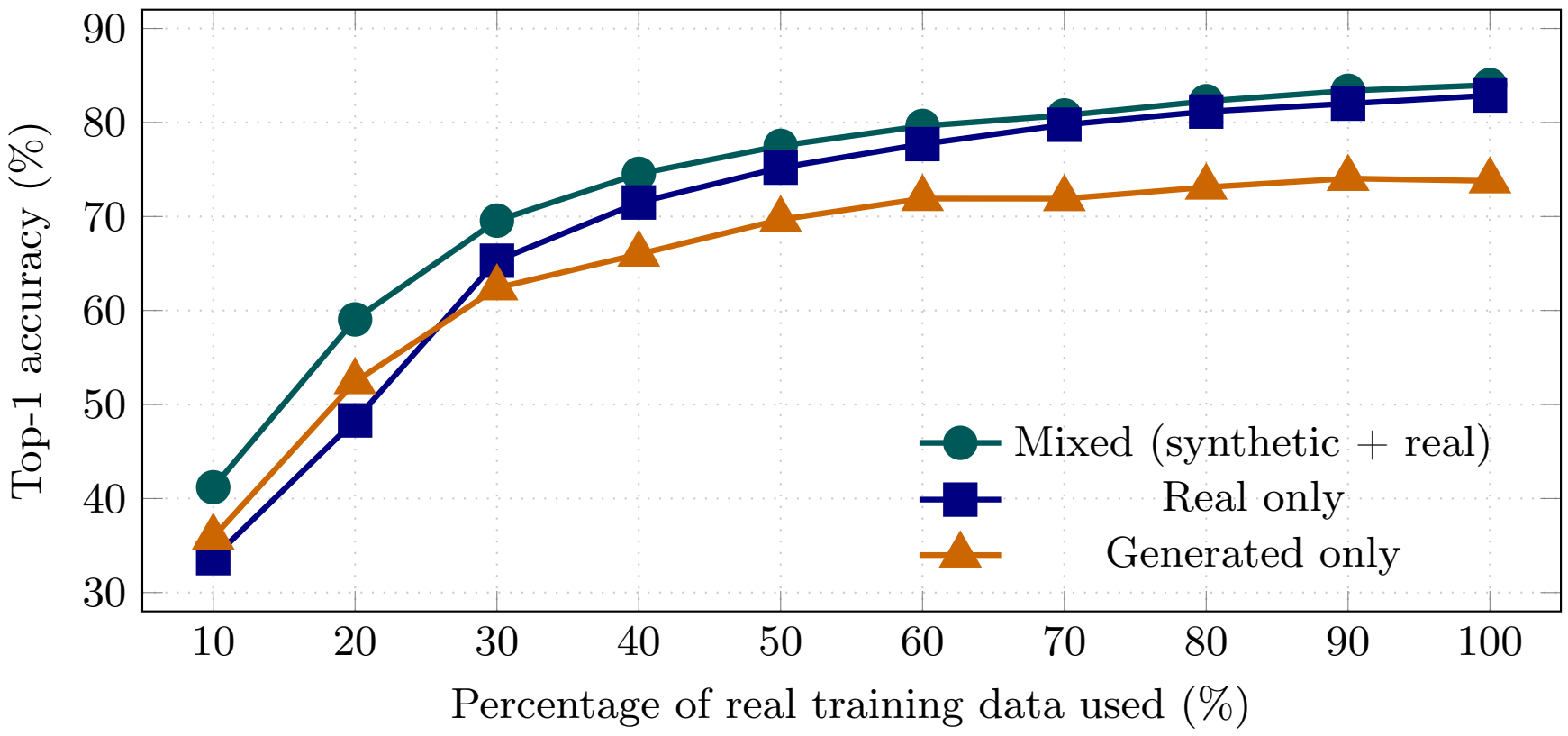
\begin{figure}[t]
    \centering
    \input{data_fraction_plot}
    \caption{Top-1 accuracy (binarised) on CVL as the fraction of real training data varies. The volume of generated data is fixed at 100\% of the full training size.}
    \label{fig:data_scarcity}
\end{figure}

Lastly, we show the effect of having less real data available during training. Fig.~\ref{fig:data_scarcity} illustrates the effect of reducing the amount of real training data used to train the diffusion and \ac{WI} model while keeping the generated data volume fixed. Firstly, we observe that the mixed scenario always outperforms the real only baseline except for 60\%, the gap is the biggest for 20\% of the training data (28\% accuracy) and smallest for 100\% (0.1\%). In low data scenarios, even the generated only baseline performs better. We therefore conclude that the synthetic data can successfully replace the real data, or can even be added to improve the \ac{WI} performance.

\subsection{Historical Handwriting Generation: BullingerDB}

We also provide baseline experiments for WI on the BullingerDB \cite{bullinger}, shown in Table \ref{tab:bull}. We observe that the margins of mixed and real only are smaller (2-3\% accuracy), and similarly to contemporary handwriting, adding synthetic data improves performance for binarized images, making our diffusion-based approach particularly attractive for identifying writers purely on stroke patterns instead of ink or pen characteristics.

\begin{table}[t]
\centering
\caption{Top-1 accuracy (in \%) on BullingerDB WI for different image formats.}
\label{tab:bull}
\begin{tabular}{lccc}
\toprule
 GR-RNN & Syn. only & Real Only & Syn. + Real \\
\midrule
Grayscale & 25.72 & \textbf{88.04} & 85.80\\
Binarised &44.83 & 77.29 & 78.62 \\

\bottomrule
\end{tabular}
\end{table}

% ---------------------------------------------------------------
\section{Conclusion} \label{sec:conclusion}
% ---------------------------------------------------------------

This paper investigated whether diffusion-generated handwriting images can be used as training data for writer identification. The results show that synthetic word images can improve writer identification accuracy when added to real data, but the benefit depends on the image format used during generation and identification.

For the CVL dataset, synthetic data generated by an RGB-trained DiffusionPen model consistently improves GR-RNN performance when combined with real data for binarized data. Increasing the volume of synthetic data further improves both Top-1 and Top-5 accuracy. In low data scenarios, adding synthetic data also helps improve the performance. Our results on BullingerDB showed that diffusion-based \ac{HTG} is a viable alternative, where the results are similarly to the CVL dataset.

Our future work will study additional evaluation for synthetic data (e.g., style interpolation to generate new writers), and a broader investigation of diffusion-based models for \ac{HTG} applied to \ac{WI}.

% ---------------------------------------------------------------
\section*{Acknowledgments}
This work was supported by the Swiss National Science Foundation (SNSF) under project grant no. 10002280, ``AI-Supported Human-Computer Interaction with Large Volumes of Historical Documents.''

% ---------------------------------------------------------------
% References
% ---------------------------------------------------------------
\bibliographystyle{splncs04}
\bibliography{bibliography}
\end{document}

%% file: visual_abstract.tex
\begin{tikzpicture}[
    box/.style={
        rectangle, rounded corners, draw=black,
        fill=gray!20,
        minimum width=2cm,
        minimum height=1.1cm,
        align=center
    },
    img/.style={
        align=center
    },
    arr/.style={-{Latex[length=3mm]}, thick}
]

% -------------------------
% (a) Generation pipeline
% -------------------------

\node[font=\sffamily\bfseries] (a_label) {(a) Handwriting Generation};
\node[below=0.4cm of a_label] (ph) {};
\node[img, left=0.9cm of ph] (realword)
{\includegraphics[height=0.05\textwidth]{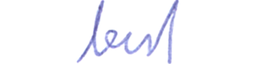}};

\node[img, right=0.9cm of ph] (text)
{\fontfamily{qcr}\selectfont calculation};

\node[box, below=0.9cm of realword] (styleenc)
{Style Encoder};

\node[box, below=1.0cm of styleenc, xshift=2.1cm] (diffusion)
{\faPen \ DiffusionPen};

\node[img, below=1.0cm of diffusion] (generated)
{\includegraphics[height=0.05\textwidth]{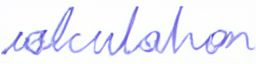}};

\draw[arr] (realword) -- (styleenc);
\draw[arr] (styleenc) |- (diffusion.west);
\draw[arr] (text) |- (diffusion.east);
\draw[arr] (diffusion) -- (generated);

% \node[
%     fit=(a_label)(realword)(text)(styleenc)(diffusion)(generated),
%     draw=black,
%     rounded corners,
%     inner sep=12pt
% ] (subfiga) {};

\node[font=\sffamily\bfseries, right=6cm of a_label] (b_label) {(b) Classification Training};

% helper styles:
% \tikzset{
%   img/.style={inner sep=1pt},
%   box/.style={draw, rounded corners, align=center, minimum width=2.5cm, minimum height=0.9cm},
%   arr/.style={-Latex, thick}
% }

% left column: Writer A
\node[font=\sffamily\small\bfseries, below=0.55cm of b_label, xshift=-3.2cm] (wa)
{Writer A};

\node[below=0.15cm of wa] (wa_real)
{
};

\node[img, right=0.01cm of wa_real.east] (wa_real1)
{   \includegraphics[height=0.05\textwidth]{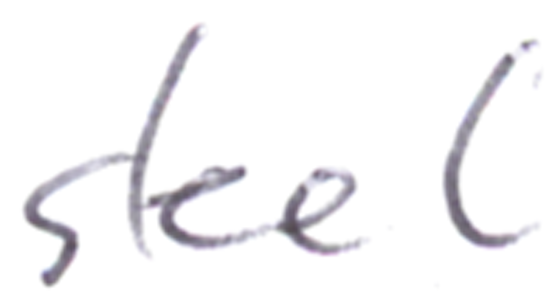}

};

\node[img, left=0.01cm of wa_real.west] (wa_real2)
{       \includegraphics[height=0.03\textwidth]{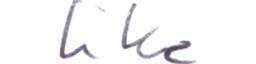}

};

\node[font=\sffamily\scriptsize, below=0.15cm of wa_real] (wa_plus)
{+ synthetic};

\node[img, below=0.08cm of wa_plus] (wa_syn)
{
    \includegraphics[height=0.03\textwidth]{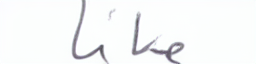}
};
\node[img, left=0.01cm of wa_syn.west] (wa_syn1){
  \includegraphics[height=0.03\textwidth]{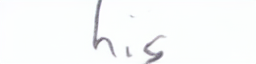}
};
\node[img, right=0.01cm of wa_syn.east] (wa_syn2){
    \includegraphics[height=0.03\textwidth]{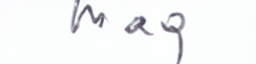}
};

% % right column: Writer B
\node[font=\sffamily\small\bfseries, right=5cm of wa] (wb)
{Writer B};

\node[below=0.15cm of wb] (wb_real)
{};
\node[img, right=0.01cm of wb_real.east] (wb_real1)
{ \includegraphics[height=0.03\textwidth]{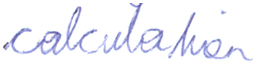}
};
\node[img, left=0.04cm of wb_real.west] (wb_real2)
{  \includegraphics[height=0.05\textwidth]{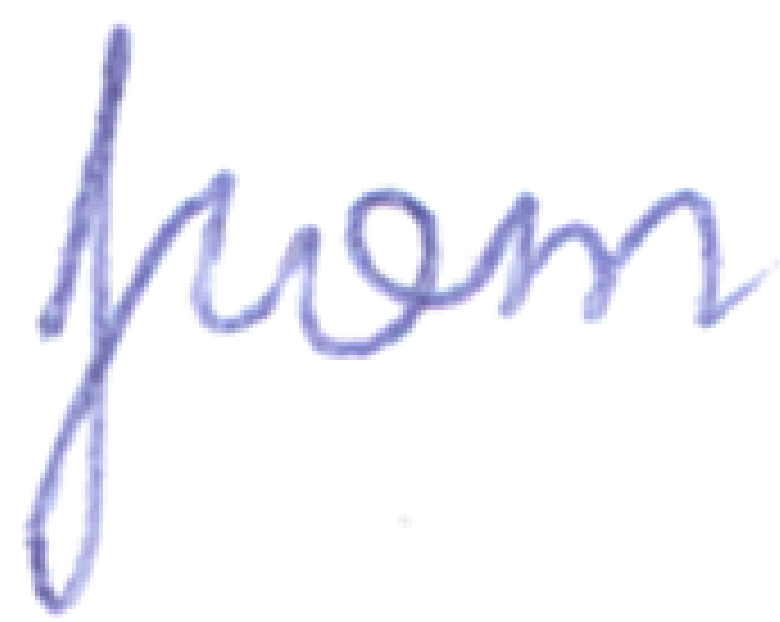}
};
\node[font=\sffamily\scriptsize, below=0.08cm of wb_real] (wb_plus)
{+ synthetic};

\node[below=0.08cm of wb_plus] (wb_syn)
{
  \includegraphics[height=0.03\textwidth]{imgs/0290-generated1-calculation.png}
};
\node[right=0.01cm of wb_syn.east] (wb_syn1)
{
  \includegraphics[height=0.03\textwidth]{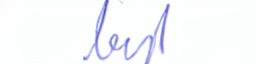}
};
\node[left=0.01cm of wb_syn.west] (wb_syn2)
{
  \includegraphics[height=0.03\textwidth]{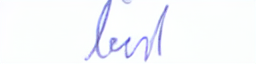}
};

% % merge point
% \node[circle, fill=black, inner sep=1.5pt, below=0.45cm of $(wa_syn)!0.5!(wb_syn)$] (merge) {};

% \draw[arr] (wa_syn.south) -- (merge);
% \draw[arr] (wb_syn.south) -- (merge);

% WI model
\node[box, below=4cm of b_label] (wi)
{Writer Identification Model};

% recognition head / person
\node[align=center, below=0.7cm of wi] (training)
{\Large\faUser\\[-1pt]\scriptsize ID};

\draw[arr] (wi) -- (training);

% % enclosing box
% \node[
%     fit=(b_label)(wa)(wa_real)(wa_plus)(wa_syn)
%         (wb)(wb_real)(wb_plus)(wb_syn1)(wi)(training),
%     draw=black,
%     rounded corners,
%     inner sep=12pt,
%     fill=gray!10
% ] (subfigb) {};
\end{tikzpicture}

%% file: imgs/generated_data_plot.tex
\begin{tikzpicture}
\begin{axis}[
    width=12cm,
    height=6cm,
    xlabel={\small Amount of generated data (\% of the real training set)},
    ylabel={\small Top-1 accuracy (\%)},
    xmin=5, xmax=520,
    ymin=34, ymax=90,
    xtick={20,50,100,200,300,400,500},
    ytick={40,50,60,70,80,90},
    grid=major,
    grid style={dotted, gray!40},
    legend pos=south east,
    legend style={draw=none, fill=none},
    tick label style={font=\small},
    label style={font=\small},
    legend style={font=\small},
]

% Baseline: Real only
\addplot[
    color=blue!50!black,
    dashed,
    very thick
] coordinates {
    (5,82.87) (520,82.87)
};
\addlegendentry{Real only (82.87)}

% Real + generated
\addplot[
    color=teal!70!black,
    mark=*,
    mark size=3pt,
    very thick
] coordinates {
    (10,82.97)
    (20,83.48)
    (50,83.56)
    (100,83.98)
    (200,84.07)
    (300,83.07)
    (400,82.14)
    (500,81.53)
};
\addlegendentry{Real + generated}

% Generated only
\addplot[
    color=orange!80!black,
    mark=triangle*,
    mark size=4pt,
    very thick
] coordinates {
    (10,37.63)
    (20,52.07)
    (50,66.83)
    (100,73.78)
    (200,78.99)
    (300,76.91)
    (400,76.41)
    (500,75.94)
};
\addlegendentry{Generated only}

% Value labels: Real + generated
\node[teal!70!black, font=\tiny, anchor=south] at (axis cs:20,84.97) {82.97};
\node[teal!70!black, font=\tiny, anchor=south] at (axis cs:30,77.48) {83.48};
\node[teal!70!black, font=\tiny, anchor=south] at (axis cs:50,85.56) {83.56};
\node[teal!70!black, font=\tiny, anchor=south] at (axis cs:100,83.98) {83.98};
\node[teal!70!black, font=\tiny, anchor=south] at (axis cs:200,85.07) {84.07};
\node[teal!70!black, font=\tiny, anchor=south] at (axis cs:300,84.07) {83.07};
\node[teal!70!black, font=\tiny, anchor=south] at (axis cs:400,83.14) {82.14};
\node[teal!70!black, font=\tiny, anchor=south] at (axis cs:500,82.53) {81.53};

% Value labels: Generated only
\node[orange!80!black, font=\tiny, anchor=north] at (axis cs:35,38.63) {37.63};
\node[orange!80!black, font=\tiny, anchor=north] at (axis cs:40,52.07) {52.07};
\node[orange!80!black, font=\tiny, anchor=north] at (axis cs:70,66.83) {66.83};
\node[orange!80!black, font=\tiny, anchor=north] at (axis cs:110,73.78) {73.78};
\node[orange!80!black, font=\tiny, anchor=north] at (axis cs:210,78.99) {78.99};
\node[orange!80!black, font=\tiny, anchor=north] at (axis cs:310,76.91) {76.91};
\node[orange!80!black, font=\tiny, anchor=north] at (axis cs:410,76.41) {76.41};
\node[orange!80!black, font=\tiny, anchor=north] at (axis cs:490,75.94) {75.94};

% Baseline label

\end{axis}
\end{tikzpicture}

%% file: data_fraction_plot.tex
\begin{tikzpicture}
\begin{axis}[
    width=12cm,
    height=6cm,
    xlabel={Percentage of real training data used (\%)},
    ylabel={Top-1 accuracy (\%)},
    xmin=5, xmax=105,
    ymin=28, ymax=92,
    xtick={10,20,30,40,50,60,70,80,90,100},
    ytick={30,40,50,60,70,80,90},
    grid=major,
    grid style={dotted, gray!40},
    legend pos=south east,
    legend style={draw=none, fill=none},
    tick label style={font=\small},
    label style={font=\small},
    legend style={font=\small},
]

% Mixed: synthetic + real
\addplot[
    color=teal!70!black,
    mark=*,
    mark size=3pt,
    very thick
] coordinates {
    (10,41.20)
    (20,59.05)
    (30,69.56)
    (40,74.54)
    (50,77.52)
    (60,79.63)
    (70,80.75)
    (80,82.26)
    (90,83.37)
    (100,83.98)
};
\addlegendentry{Mixed (synthetic + real)}

% Real only
\addplot[
    color=blue!50!black,
    mark=square*,
    mark size=3pt,
    very thick
] coordinates {
    (10,33.66)
    (20,48.31)
    (30,65.30)
    (40,71.49)
    (50,75.19)
    (60,77.70)
    (70,79.76)
    (80,81.16)
    (90,82.00)
    (100,82.87)
};
\addlegendentry{Real only}

% Generated only
\addplot[
    color=orange!80!black,
    mark=triangle*,
    mark size=4pt,
    very thick
] coordinates {
    (10,35.88)
    (20,52.42)
    (30,62.40)
    (40,65.99)
    (50,69.67)
    (60,71.90)
    (70,71.89)
    (80,73.11)
    (90,74.04)
    (100,73.78)
};
\addlegendentry{Generated only}

\end{axis}
\end{tikzpicture}

%% file: bibliography.bib
@inproceedings{cvl,
  author    = {Florian Kleber and
               Stefan Fiel and
               Markus Diem and
               Robert Sablatnig},
  title     = {{CVL-DataBase}: An Off-Line Database for Writer Retrieval, Writer Identification
               and Word Spotting},
  booktitle = {
               {ICDAR} 2013},
  pages     = {560--564},
  year      = {2013},
}

@inproceedings{icdar13,
  author    = {Georgios Louloudis and
               Basilios Gatos and
               Nikolaos Stamatopoulos and
               A. Papandreou},
  title     = {{ICDAR} 2013 Competition on Writer Identification},
  booktitle = {12th International Conference on Document Analysis and Recognition,
               {ICDAR} 2013, Washington, DC, USA, August 25-28, 2013},
  pages     = {1397--1401},
  year      = {2013}
}

@inproceedings{peer_saghog_2024,
    address={Athens, Greece},
	title = {{SAGHOG}: {S}elf-{S}upervised {A}utoencoder for {G}enerating {HOG} {F}eatures for {W}riter {R}etrieval},
	booktitle = {{ICDAR 2024}},
	author = {Peer, Marco and Kleber, Florian and Sablatnig, Robert},
	year = {2024},
	pages = {121--138},
}

@article{gr-rnn,
  author       = {Sheng He and
                  Lambert Schomaker},
  title        = {{GR-RNN:} Global-context residual recurrent neural networks for writer
                  identification},
  journal      = {Pattern Recognit.},
  volume       = {117},
  pages        = {107975},
  year         = {2021},
}

@inproceedings{diffpen,
  author       = {Konstantina Nikolaidou and
                  George Retsinas and
                  Giorgos Sfikas and
                  Marcus Liwicki},
  title        = {{DiffusionPen: Towards Controlling the Style of Handwritten Text Generation}},
  booktitle    = {{ECCV} 2024},
  pages        = {417--434},
  year         = {2024},
}

@inproceedings{bullinger,
  author       = {Marco Peer and Anna Scius-Bertrand and Patricia Scheurer and Andreas Fischer},
  title        = {{BullingerDB: A Dataset for Handwritten Text Recognition and Writer Retrieval}},
  booktitle    = {ICDAR 2026 (accepted)},
  year         = {2026},
}

@inproceedings{wordstylist,
  author       = {Konstantina Nikolaidou and
                  George Retsinas and
                  Vincent Christlein and
                  Mathias Seuret and
                  Giorgos Sfikas and
                  Elisa Barney Smith and
                  Hamam Mokayed and
                  Marcus Liwicki},
  title        = {{WordStylist: Styled Verbatim Handwritten Text Generation with Latent
                  Diffusion Models}},
  booktitle    = {{ICDAR} 2023},
  pages        = {384--401},
  year         = {2023},
}

@inproceedings{muth2024,
  author       = {Markus Muth and
                  Marco Peer and
                  Florian Kleber and
                  Robert Sablatnig},
  title        = {{Maximizing Data Efficiency of {HTR} Models by Synthetic Text}},
  booktitle    = {{DAS}
                  2024},
  pages        = {295--311},
  year         = {2024},
}

@article{fragnet,
  author       = {Sheng He and
                  Lambert Schomaker},
  title        = {FragNet: Writer Identification Using Deep Fragment Networks},
  journal      = {{IEEE} Trans. Inf. Forensics Secur.},
  volume       = {15},
  pages        = {3013--3022},
  year         = {2020},
}

@inproceedings{gorges2026circleid,
  title={{ICDAR 2026 Competition on Writer Identification and Pen Classification from Hand-Drawn Circles}},
  author={Gorges, Thomas and van der Loop, Janne and H{\"u}ttner, Lukas and Schneider, Linda-Sophie and Wu, Fei and Seuret, Mathias and Christlein, Vincent},
  journal={arXiv preprint arXiv:2605.07816},
  year={2026},
  booktitle    = {ICDAR 2026 (accepted)},
}

@inproceedings{koepf,
  author       = {Michael Koepf and
                  Florian Kleber and
                  Robert Sablatnig},
  title        = {Writer Identification and Writer Retrieval Using Vision Transformer
                  for Forensic Documents},
  booktitle    = { {DAS}
                  2022},
  pages        = {352--366},
  year         = {2022},

}

@article{mayr,
  author       = {Martin Mayr and
                  Marcel Dreier and
                  Florian Kordon and
                  Mathias Seuret and
                  Jochen Z{\"{o}}llner and
                  Fei Wu and
                  Andreas Maier and
                  Vincent Christlein},
  title        = {Zero-Shot Paragraph-level Handwriting Imitation with Latent Diffusion
                  Models},
  journal      = {Int. J. Comput. Vis.},
  volume       = {133},
  number       = {10},
  pages        = {7054--7075},
  year         = {2025},
}

@inproceedings{scrabblegan,
  author       = {Sharon Fogel and
                  Hadar Averbuch{-}Elor and
                  Sarel Cohen and
                  Shai Mazor and
                  Roee Litman},
  title        = {ScrabbleGAN: Semi-Supervised Varying Length Handwritten Text Generation},
  booktitle    = {
                  {CVPR} 2020},
  pages        = {4323--4332},
  year         = {2020},
}

@inproceedings{GANwriting,
  author       = {Lei Kang and
                  Pau Riba and
                  Yaxing Wang and
                  Mar{\c{c}}al Rusi{\~{n}}ol and
                  Alicia Forn{\'{e}}s and
                  Mauricio Villegas},
  title        = {GANwriting: Content-Conditioned Generation of Styled Handwritten Word
                  Images},
  booktitle    = {{ECCV} 2020},
  pages        = {273--289},
  year         = {2020},
}
